\documentclass{article}

\usepackage[final]{tackling_climate_workshop_style}

\usepackage[utf8]{inputenc} 
\usepackage[T1]{fontenc}    
\usepackage{hyperref}       
\usepackage{url}            
\usepackage{booktabs}       
\usepackage{amsfonts}       
\usepackage{nicefrac}       
\usepackage{microtype}      

\usepackage{graphicx}       

\title{\textit{DemandCast}: Global hourly electricity demand forecasting}

\author{
  Kevin Steijn\\
  Open Energy Transition\\
  Bayreuth, Germany\\
  \texttt{kevin.steijn@openenergytransition.org}\\
  \And
  Vamsi Priya Goli\\
  Open Energy Transition\\
  Bayreuth, Germany\\
  \texttt{goli.vamsi@openenergytransition.org}\\
  \And
  Enrico Antonini\\
  Open Energy Transition\\
  Bayreuth, Germany\\
  \texttt{enrico.antonini@openenergytransition.org}
}

\begin{document}

\maketitle

\begin{abstract}
  This paper presents a machine learning framework for electricity demand forecasting across diverse geographical regions using the gradient boosting algorithm XGBoost. The model integrates historical electricity demand and comprehensive weather and socioeconomic variables to predict normalized electricity demand profiles. To enable robust training and evaluation, we developed a large-scale dataset spanning multiple years and countries, applying a temporal data-splitting strategy that ensures benchmarking of out-of-sample performance. Our approach delivers accurate and scalable demand forecasts, providing valuable insights for energy system planners and policymakers as they navigate the challenges of the global energy transition.
\end{abstract}

\section{Introduction}

Scientific evidence highlights the importance of limiting global warming driven by anthropogenic carbon emissions to manageable levels to avoid disruptive impacts on natural and human systems \cite{meinshausen2009greenhouse}. Addressing this challenge requires broad transformations, with the decarbonization of electricity systems—which currently account for approximately 33\% of global carbon emissions \cite{owid-co2-and-greenhouse-gas-emissions}—playing a pivotal role in reducing them. The transition toward low-carbon electricity involves shifting generation away from fossil fuels toward sources such as wind, solar, hydropower and nuclear \cite{sepulveda2018role}. Equally critical to this transition is the decarbonization of energy end-uses—namely transportation, heating, and industry—that together account for approximately 28\% of global emissions \cite{owid-co2-and-greenhouse-gas-emissions}. Electrifying some or all of these sectors presents a major opportunity to reduce carbon emissions, provided that the electricity originates from low-carbon sources.

In this context, there is considerable uncertainty regarding future energy and electricity consumption patterns, as these will be shaped by dynamic socio-economic factors such as population growth, economic development, urbanization, technological changes, and evolving standards of living \cite{bossmann2015shape}. This transformation is especially critical in countries across the Global South, where expanding access to reliable electricity is crucial for economic and social development. These countries face the challenge of meeting growing energy demand while following pathways that minimize carbon emissions \cite{duan2020balancing}. These uncertainties make it challenging to anticipate precisely how much, where, and when electricity will be demanded in the coming years.

To address these uncertainties, we have developed DemandCast \cite{demandcast}, a model to support energy system planners and policymakers. It is an open-source machine learning framework designed to produce synthetic hourly electricity demand forecasts for countries with limited historical data or looking to explore future scenarios. It is trained on historical hourly or sub-hourly electricity demand, socioeconomic and weather data using a gradient boosting algorithm. DemandCast helps improve the accuracy of demand forecasting, which is essential for integrating variable renewable resources and managing electricity grids effectively during this period of transformation.
 
\section{Data}

The DemandCast framework integrates data from multiple sources, including historical electricity demand, weather variables, and socioeconomic indicators. Each dataset is collected, pre-processed, and harmonized through modular data pipelines that facilitate the inclusion of additional countries and data sources while ensuring reproducibility and transparency.

\begin{figure}[ht]
    \centering
    \includegraphics[width=0.8\textwidth]{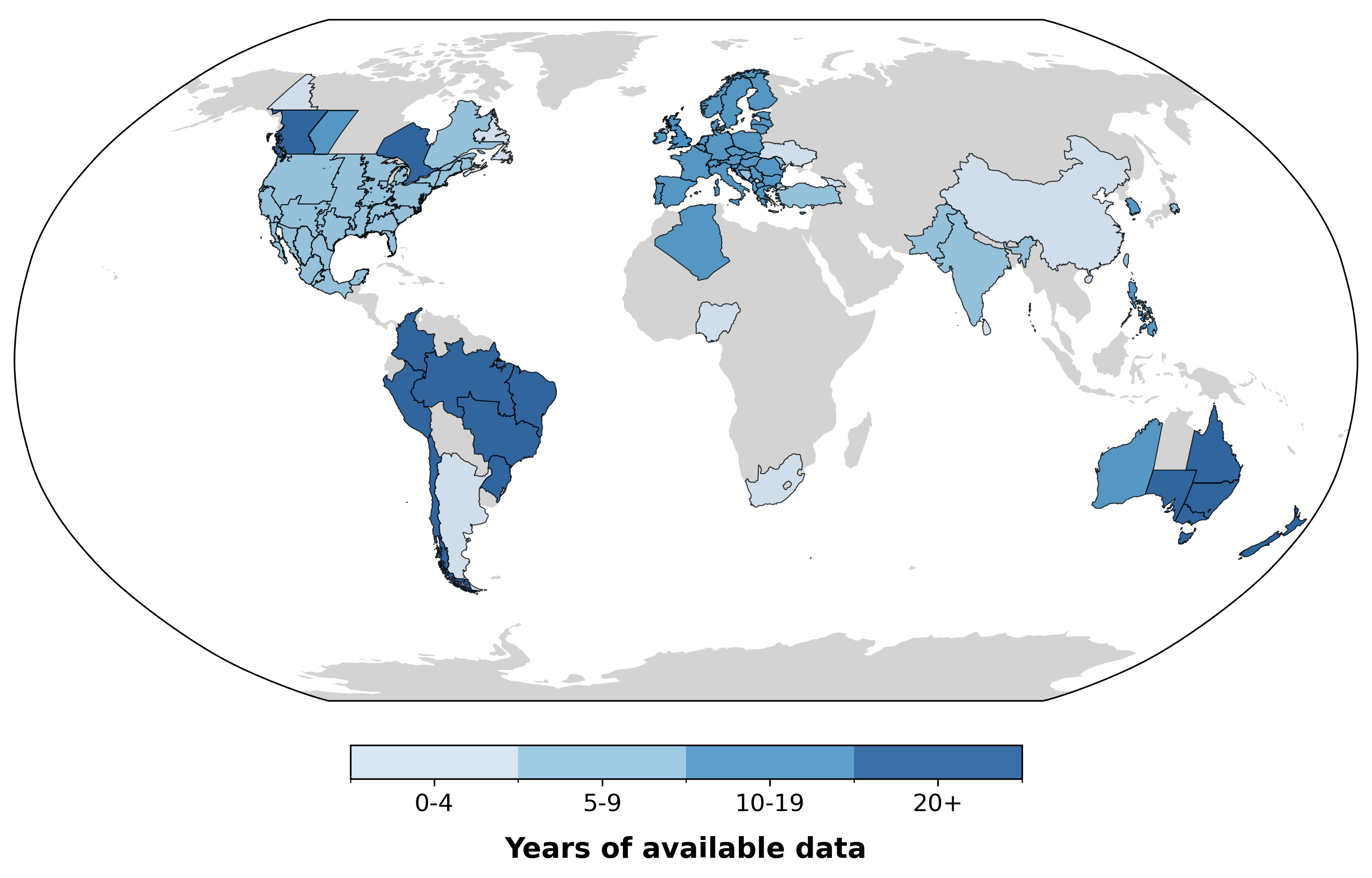}
    \caption{Countries and subdivisions for which retrieval modules of electricity demand data are available.}
    \label{fig:data_availability_map}
\end{figure}

A major obstacle to developing robust machine learning frameworks for demand forecasting is the limited availability of high-quality, high-resolution electricity demand data across many regions, which restricts both the training of accurate models and their application to long-term scenario analysis. To address this gap, we developed the openly accessible Awesome Electricity Demand repository \cite{aed}, which systematically compiles links to public hourly and sub-hourly datasets. The repository is designed to improve the accessibility and discoverability of demand data for energy system modelers and planners. Within the DemandCast framework, retrieval modules access this resource and harmonize all time series to a unified structure, standardizing timestamps to UTC and demand values to megawatts (MW). Figure \ref{fig:data_availability_map} presents the countries and subdivisions currently covered, along with the temporal availability of demand data.

Temperature is incorporated as a key predictor, as electricity demand is highly sensitive to climatic conditions in regions with widespread use of cooling and heating appliances such as air conditioners and heat pumps. To represent this effect, gridded hourly temperature records are retrieved from the Copernicus ERA5 reanalysis dataset \cite{era5}. Features are derived from the top one and top three most populous grid cells, with additional monthly averages and rankings constructed to capture seasonal variation.

Beyond demand and temperature, DemandCast retrieves auxiliary variables to account for the broader socioeconomic drivers of electricity consumption. These include gridded population data from the Socioeconomic Data and Applications Center \cite{ciesin2018gpwv4}, gridded gross domestic product per capita from global economic databases \cite{wang2022global}, and country-level annual electricity demand per capita estimates \cite{ember_yearly_electricity_data_2025}.

\section{Model}

DemandCast builds on recent developments in electricity demand forecasting using machine learning methods. Gradient boosting algorithms such as XGBoost \cite{chen2016xgboost} have demonstrated strong potential for predicting high-resolution national demand \cite{mattsson2021autopilot, gegis}. Extending these advances, DemandCast expands the temporal coverage, geographic scope, and transparency of existing approaches by providing a fully reproducible, end-to-end pipeline. While prior studies relied on demand data from a single year (2015) covering 44 countries, DemandCast incorporates data spanning 2000 to 2025 (where available) and increases geographic representation to 56 countries, including subdivisions for large nations such as the United States, Canada, Mexico, Brazil, and Australia.

We introduce a normalized target variable, $D_n$, computed in each calendar year, which represents hourly electricity demand as a fraction of annual demand, adjusted for data coverage:

\begin{equation}
D_n(t) = \frac{D(t)}{D_Y} \frac{\sum_Y H_{avail}}{\sum_Y H},
\end{equation}

\noindent where $D(t)$ is the electricity demand in MW at time $t$, $D_Y$ is the total electricity demand in MWh for the year $Y$, $\sum_Y H_{avail}$ is the number of hours available in the year $Y$, and $\sum_Y H$ is the total number of hours in the year $Y$.

This formulation ensures that the machine learning model predicts the temporal profile of electricity demand rather than its absolute magnitude. By normalizing demand to annual consumption, the model can capture short- and long-term variations such as daily, weekly, or seasonal patterns. Absolute demand levels are then reconstructed in a post-processing step by scaling the predicted temporal pattern with the corresponding estimates of annual demand. This approach also compensates for incomplete datasets by adjusting for the proportion of hours that are actually observed.

\begin{figure}[ht]
    \centering
    \includegraphics[width=0.6\textwidth]{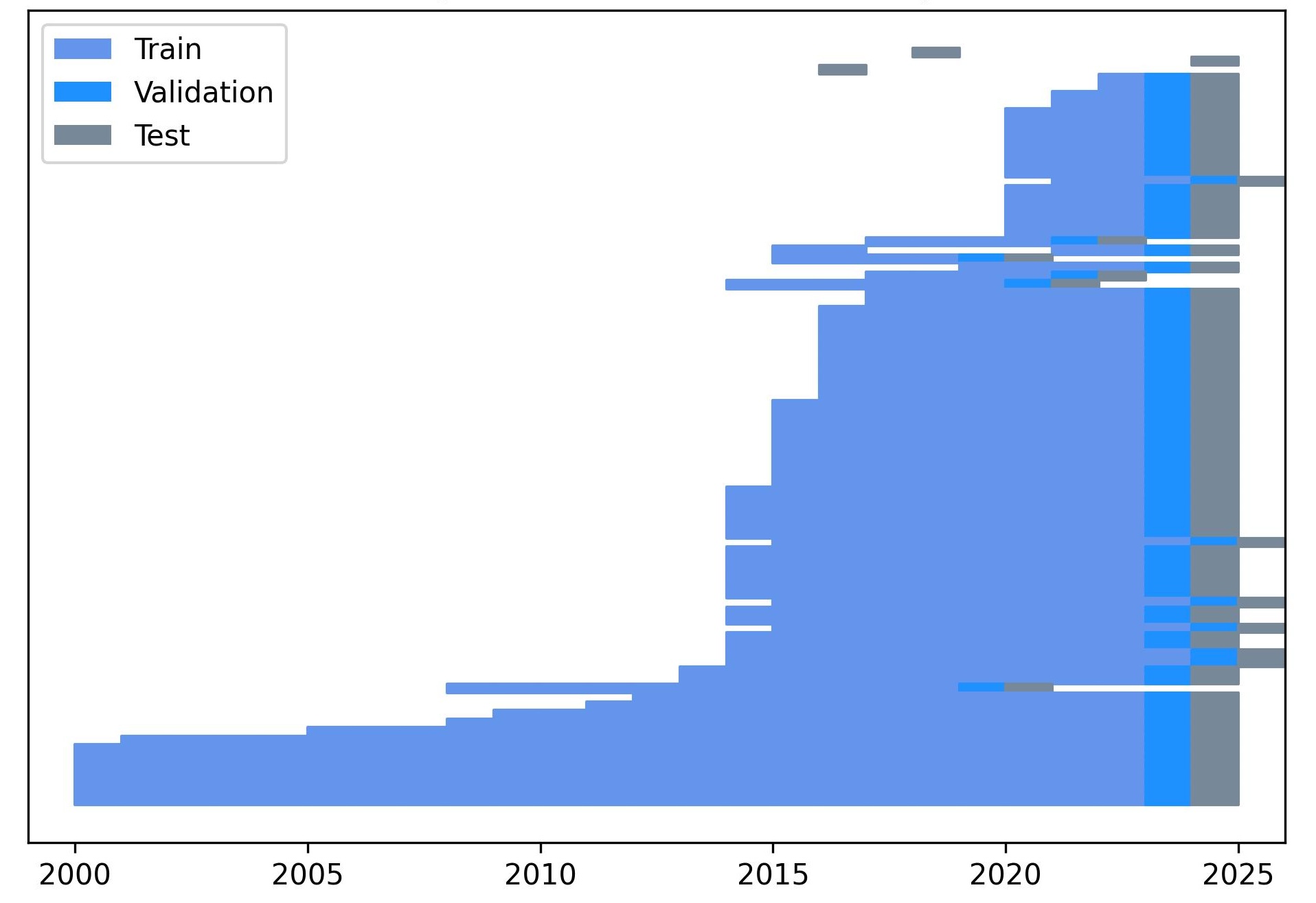}
    \caption{Training, test, and validation sets in which available electricity demand was split.}
    \label{fig:data_available}
\end{figure}

Model training follows a data-splitting strategy, as shown in Fig. \ref{fig:data_available}. For each region, the final available year of demand data is reserved for testing (8.91\% of the dataset), while the second-to-last year is used for validation (9.84\%). All remaining years form the training dataset (81.25\%). This structure ensures that model evaluation is based on out-of-sample years, thereby providing a realistic assessment of generalization to future demand patterns.

\section{Results and Discussion}

The XGBoost model was trained on 6,041,222 hourly electricity demand observations, with an additional 731,538 observations used for validation and 662,369 observations for testing. Across all regions in the test set, the model achieved an average mean absolute percentage error (MAPE) of 9.2\%. This result is comparable to previous studies, which reported a MAPE of around 8\%, albeit using smaller training datasets and less diverse regions.

\begin{figure}[ht]
    \centering
    \includegraphics[width=\textwidth]{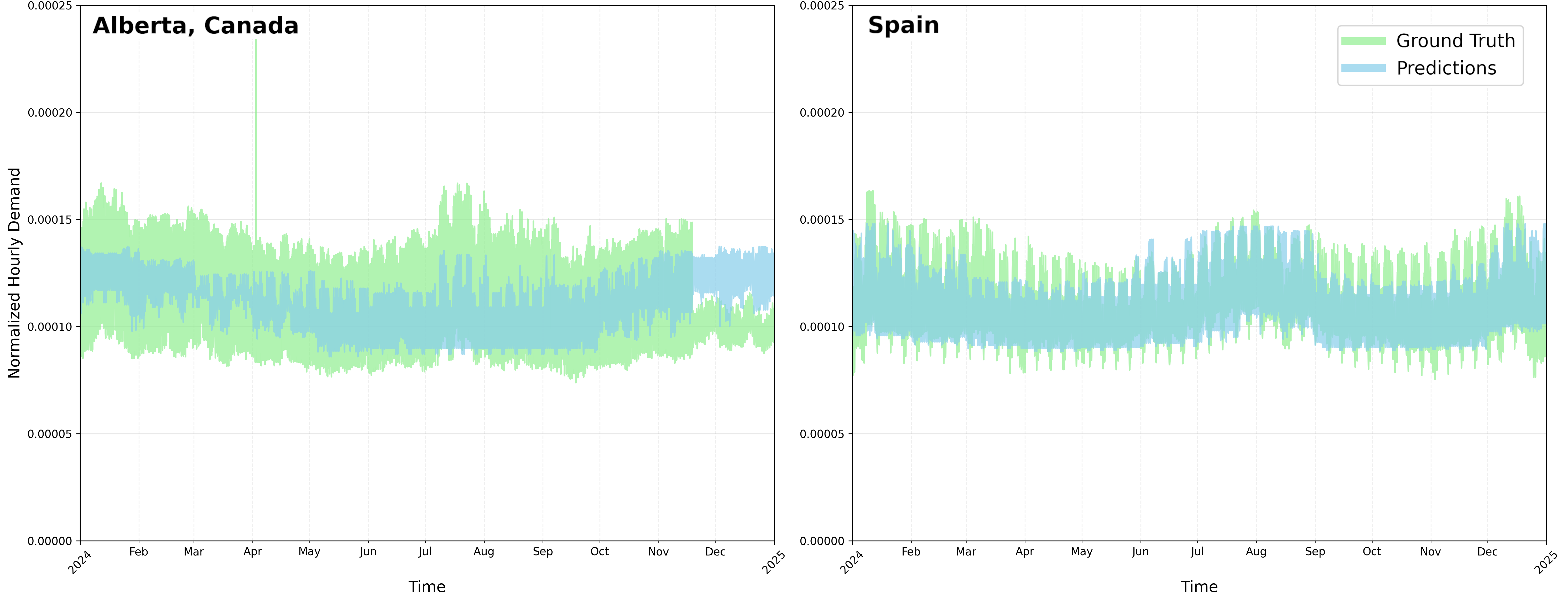}
    \caption{Comparison between historical and forecast electricity demand.}
    \label{fig:normalized_profiles}
\end{figure}

To provide a qualitative performance assessment, Fig. \ref{fig:normalized_profiles} illustrates the normalized demand profiles of the best and worst performing regions. This highlights both the model’s ability to reproduce temporal demand patterns and the variability in prediction accuracy across different regions. A complete view of the regional MAPE values is provided in Fig. \ref{fig:MAPE} in the Appendix, which also reports all numerical MAPE results for the training, validation, and testing sets.

The characteristics of the training data exert an important influence on the robustness of these results. The temporal data-splitting strategy, which reserves the most recent full year for testing and the preceding year for validation, ensures chronological integrity but also highlights imbalances arising from heterogeneous data availability across regions. Some countries contribute extensive multi-year training histories, while others are represented by relatively short time spans, which may lead to uneven geographic performance. Expanding the scope of historical demand collection will help alleviate this imbalance, while alternative training strategies, such as clustering countries by structural similarities, may further improve model performance in under-represented regions.

While effective as a forecasting tool, the current model architecture lacks the ability to quantify predictive uncertainty. This limitation is particularly relevant for long-term planning applications, where decision-making requires an explicit consideration of uncertainty. Future work will therefore explore alternative architectures capable of generating probabilistic forecasts, as well as the potential to enhance predictive performance through systematic hyperparameter optimization of the gradient boosting algorithm. Further experiments with other gradient boosting implementations may also be conducted as benchmarks against XGBoost.

\section{Conclusion}
This study presents DemandCast, an open-source machine learning framework designed for forecasting global hourly electricity demand. By integrating electricity demand, weather, and socioeconomic datasets with a gradient boosting model, DemandCast advances previous approaches through expanded temporal coverage (2000–2025), wider geographic scope (56 countries), and a fully reproducible end-to-end pipeline. The model effectively captures temporal demand patterns, achieving an average MAPE of 9.2\% on out-of-sample data, while also revealing challenges related to heterogeneous data availability. Ongoing efforts in data collection, feature refinement, and methodological improvements—including clustering, hyperparameter optimization, and probabilistic forecasting—are expected to further enhance the model’s generalizability. By providing a scalable and transparent tool, DemandCast supports energy planners, policymakers, and researchers in navigating uncertainties surrounding future electricity demand amid the global energy transition.

\bibliographystyle{unsrt}
\bibliography{references}

\section*{Appendix}

\begin{figure}[ht]
    \centering
    \includegraphics[width=\textwidth]{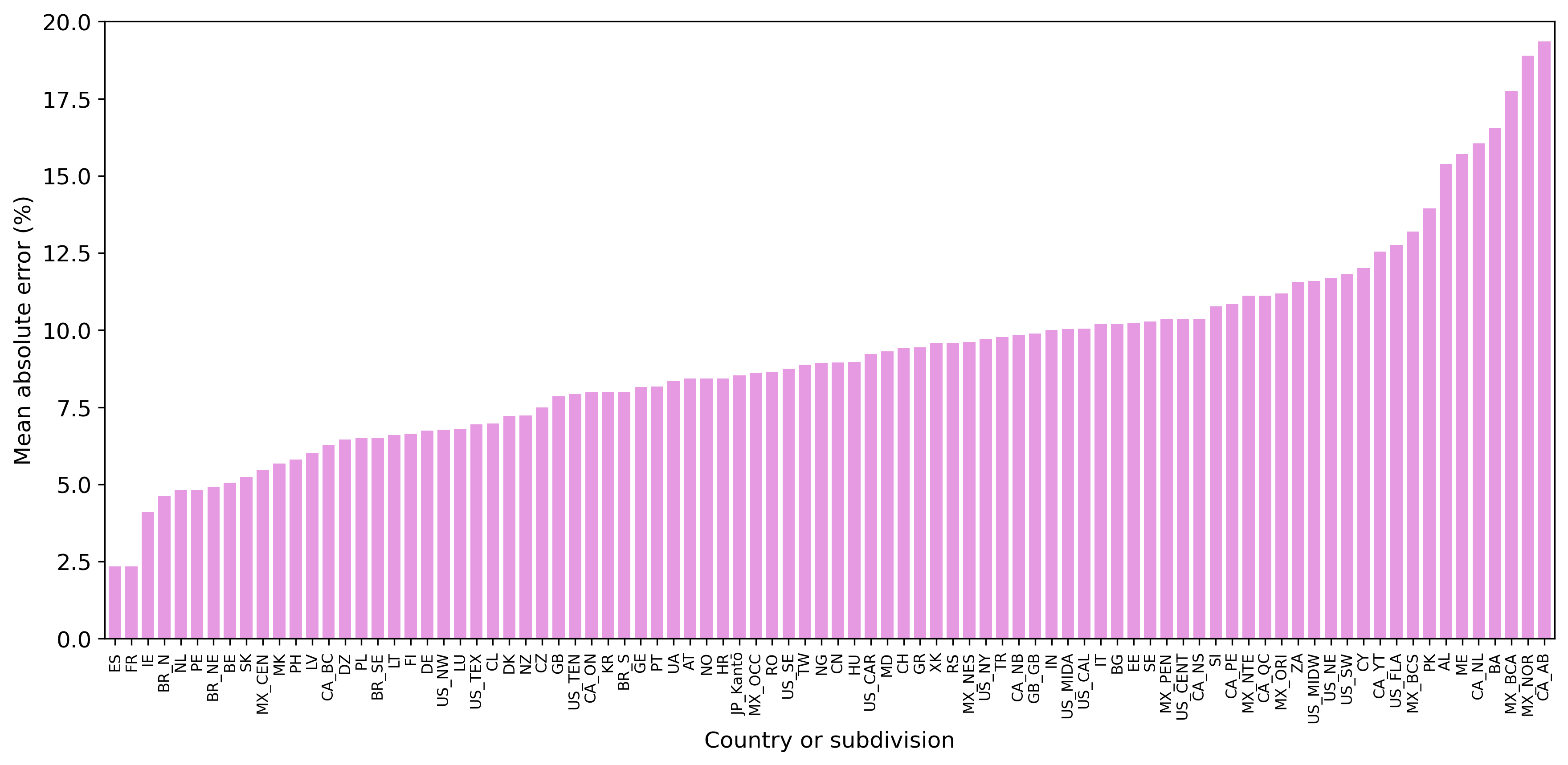}
    \caption{MAPE values resulting from the testing of DemandCast.}
    \label{fig:MAPE}
\end{figure}

\begin{table}[ht]
\begin{tabular}{llll}
entity\_code & MAPE\_train & MAPE\_val & MAPE\_test \\
AL           & 0.2078      & 0.4397    & 0.1539     \\
AT           & 0.0787      & 0.0815    & 0.0843     \\
BA           & 0.1093      & 0.1273    & 0.1656     \\
BE           & 0.0539      & 0.0546    & 0.0505     \\
BG           & 0.0783      & 0.1106    & 0.1020     \\
BR\_N        & 0.0512      & 0.0492    & 0.0462     \\
BR\_NE       & 0.0546      & 0.0580    & 0.0493     \\
BR\_S        & 0.0860      & 0.0905    & 0.0800     \\
BR\_SE       & 0.0650      & 0.0714    & 0.0652     \\
CA\_AB       & 0.0549      & 0.2056    & 0.1936     \\
CA\_BC       & 0.0553      & 0.0506    & 0.0628     \\
CA\_NB       & 0.1000      & 0.0887    & 0.0985     \\
CA\_NL       & 0.1347      & 0.1741    & 0.1605     \\
CA\_NS       & 0.0881      & 0.0966    & 0.1037     \\
CA\_ON       & 0.0709      & 0.0757    & 0.0798     \\
CA\_PE       & 0.0683      & 0.0789    & 0.1085     \\
CA\_QC       & 0.1155      & 0.1097    & 0.1112     \\
CA\_YT       &             &           & 0.1254     \\
CH           & 0.0752      & 0.0885    & 0.0941     \\
CL           & 0.0727      & 0.0648    & 0.0698     \\
CN           &             &           & 0.0895     \\
CY           & 0.1129      & 0.1067    & 0.1201     \\
CZ           & 0.0623      & 0.0714    & 0.0749     \\
DE           & 0.0713      & 0.0667    & 0.0674     \\
DK           & 0.0679      & 0.0646    & 0.0722     \\
DZ           & 0.0897      & 0.0874    & 0.0646     \\
EE           & 0.0692      & 0.0747    & 0.1023     \\
ES           & 0.0634      & 0.0622    & 0.0234     \\
FI           & 0.0548      & 0.0624    & 0.0664     \\
FR           & 0.1156      & 0.1055    & 0.0234     \\
GB           & 0.0765      & 0.1067    & 0.0786     \\
GB\_GB       & 0.0790      & 0.0922    & 0.0989     \\
GE           & 0.0679      & 0.0746    & 0.0816     \\
GR           & 0.0786      & 0.0870    & 0.0945     \\
HR           & 0.0693      & 0.0696    & 0.0843     \\
HU           & 0.0576      & 0.0793    & 0.0896
\end{tabular}
\label{tab:MAPE_values_1}
\end{table}

\begin{table}[ht]
\begin{tabular}{llll}
entity\_code & MAPE\_train & MAPE\_val & MAPE\_test \\
IE           & 0.0562      & 0.0418    & 0.0411     \\
IN           & 0.0815      & 0.0701    & 0.1001     \\
IT           & 0.1085      & 0.1049    & 0.1019     \\
JP\_Kantō    & 0.0717      & 0.0821    & 0.0853     \\
KR           & 0.0586      & 0.0661    & 0.0800     \\
LT           & 0.0701      & 0.0597    & 0.0660     \\
LU           & 0.0838      & 0.0664    & 0.0680     \\
LV           & 0.0752      & 0.0653    & 0.0603     \\
MD           & 0.0900      & 0.0829    & 0.0931     \\
ME           & 0.0997      & 0.1484    & 0.1570     \\
MK           & 0.1395      & 0.2952    & 0.0567     \\
MX\_BCA      & 0.1434      & 0.1552    & 0.1775     \\
MX\_BCS      & 0.1065      & 0.1434    & 0.1319     \\
MX\_CEN      & 0.1151      & 0.0514    & 0.0547     \\
MX\_NES      & 0.0916      & 0.0931    & 0.0962     \\
MX\_NOR      & 0.1305      & 0.1775    & 0.1890     \\
MX\_NTE      & 0.1057      & 0.1051    & 0.1111     \\
MX\_OCC      & 0.1228      & 0.0877    & 0.0862     \\
MX\_ORI      & 0.1099      & 0.1083    & 0.1119     \\
MX\_PEN      & 0.1462      & 0.0960    & 0.1035     \\
NG           &             &           & 0.0894     \\
NL           & 0.0795      & 0.0725    & 0.0481     \\
NO           & 0.0862      & 0.0848    & 0.0843     \\
NZ           & 0.0711      & 0.0741    & 0.0724     \\
PE           & 0.0525      & 0.0511    & 0.0482     \\
PH           & 0.0684      & 0.0620    & 0.0581     \\
PK           & 0.1103      & 0.1441    & 0.1394     \\
PL           & 0.0620      & 0.0612    & 0.0650     \\
PT           & 0.0718      & 0.0788    & 0.0818     \\
RO           & 0.0567      & 0.0773    & 0.0865     \\
RS           & 0.0959      & 0.0926    & 0.0959     \\
SE           & 0.0841      & 0.0808    & 0.1028     \\
SI           & 0.0737      & 0.1052    & 0.1077     \\
SK           & 0.0527      & 0.0566    & 0.0525     \\
TR           & 0.0694      & 0.0679    & 0.0978     \\
TW           & 0.0616      & 0.0613    & 0.0888     \\
UA           & 0.0611      & 0.0669    & 0.0835     \\
US\_CAL      & 0.0984      & 0.1046    & 0.1005     \\
US\_CAR      & 0.0888      & 0.0908    & 0.0922     \\
US\_CENT     & 0.0955      & 0.1073    & 0.1036     \\
US\_FLA      & 0.1326      & 0.1309    & 0.1276     \\
US\_MIDA     & 0.0985      & 0.0952    & 0.1004     \\
US\_MIDW     & 0.0991      & 0.1066    & 0.1160     \\
US\_NE       & 0.1049      & 0.1110    & 0.1170     \\
US\_NW       & 0.0637      & 0.0596    & 0.0678     \\
US\_NY       & 0.0969      & 0.0923    & 0.0972     \\
US\_SE       & 0.0918      & 0.0895    & 0.0875     \\
US\_SW       & 0.1133      & 0.1103    & 0.1181     \\
US\_TEN      & 0.0847      & 0.0794    & 0.0793     \\
US\_TEX      & 0.0764      & 0.0804    & 0.0694     \\
XK           & 0.1789      & 0.1742    & 0.0959     \\
ZA           & 0.1175      & 0.1173    & 0.1157     \\
             &             &           &            \\
Average      & 0.0878      & 0.0966    & 0.0920     \\
Min          & 0.0512      & 0.0418    & 0.0234     \\
Max          & 0.2078      & 0.4397    & 0.1936    
\end{tabular}
\label{tab:MAPE_values_2}
\end{table}

\end{document}